\documentclass[lettersize,journal]{IEEEtran}

\usepackage{xcolor}
\usepackage{float}
\usepackage{pifont}
\usepackage{booktabs} 
\usepackage{multirow}
\usepackage{xspace}
\usepackage{enumitem}
\usepackage{balance}
\usepackage[multiple]{footmisc}
\usepackage[autostyle]{csquotes}
\usepackage{listings}
\usepackage{tikz}
\usepackage{amsmath,amssymb,amsfonts}
\usepackage{algorithmic}

\usepackage{pifont}
\usepackage{graphicx}
\usepackage{textcomp}
\usepackage{adjustbox}
\usepackage{lscape}
\usepackage{tabularx}
\usepackage{color}
\usepackage[newcommands]{ragged2e}
\usepackage{colortbl}
\usepackage{array}
\usepackage{subcaption,lipsum,booktabs}
\usepackage{siunitx} 
\usepackage{hyperref}
\usepackage{xurl}

\usepackage{algorithm}

\begin{document}


\title{Federated Learning for Misbehaviour Detection with Variational Autoencoders and Gaussian Mixture Models}

\author{Enrique Mármol Campos, Aurora Gonzalez-Vidal Jos\'{e} L. Hern\'{a}ndez-Ramos and Antonio Skarmeta

\thanks{Enrique Marmol, Aurora Gonzalez-Vidal Jos\'{e} L. Hern\'{a}ndez-Ramos and Antonio Skarmeta are with the University of Murcia, Spain. E-mail: \{enrique.marmol, aurora.gonzalez2, jluis.hernandez, skarmeta\}@um.es}}

\maketitle

\begin{abstract}
Federated Learning (FL) has become an attractive approach to collaboratively train Machine Learning (ML) models while data sources’ privacy is still preserved. However, most of existing FL approaches are based on supervised techniques, which could require resource-intensive activities and human intervention to obtain labelled datasets. Furthermore, in the scope of cyberattack detection, such techniques are not able to identify previously unknown threats. In this direction, this work proposes a novel unsupervised FL approach for the identification of potential misbehavior in vehicular environments. We leverage the computing capabilities of public cloud services for model aggregation purposes, and also as a central repository of misbehavior events, enabling cross-vehicle learning and collective defense strategies. Our solution integrates the use of Gaussian Mixture Models (GMM) and Variational Autoencoders (VAE) on the VeReMi dataset in a federated environment, where each vehicle is intended to train only with its own data. Furthermore, we use Restricted Boltzmann Machines (RBM) for pre-training purposes, and Fed+ as aggregation function to enhance model’s convergence. Our approach provides better performance (more than 80\%) compared to recent proposals, which are usually based on supervised techniques and artificial divisions of the VeReMi dataset. 
\end{abstract}


\section{Introduction}\label{sec:Introduction}
The application of Machine Learning (ML) techniques for cyberattack detection has attracted significant interest in recent years \cite{ahmad2021network}. However, most of the proposed approaches are based on centralized deployments, which are intended to manage massive amounts of data from different sources. Therefore, all the data produced by such sources is typically shared through a data center. This scenario sets out several issues related to the delay associated with the centralized reasoning process that is usually carried out in the cloud, as well as privacy concerns, especially in the case of sensitive data. Indeed, recent works \cite{zhang2021survey, lyu2020threats, aledhari2020federated} state the importance of protecting clients' personal information, and the need to develop distributed ML approaches to cope with the problems of centralized systems in terms of limited communication bandwidth, intermittent network connectivity, and strict delay constraints \cite{lyu2020threats}. These aspects have motivated researchers to move toward more decentralized ML learning frameworks \cite{billah2022systematic}. In this direction, Federated learning (FL) was proposed \cite{mcmahan2017communication} as a collaborative ML approach where the different data sources (or clients) train a common model, which is updated through an aggregator entity. Therefore, data is not communicated to any external entity and clients benefit from the training of other clients without sharing any information about their dataset. The server aggregates all the weights using an aggregation function and sends the result to the clients so that they continue their training using the updated information. 

An example of a cybersecurity problem that has recently gained significant interest is the well-known \textit{misbehavior detection} in vehicular environments which usually refers to the detection of vehicles transmitting false information that cannot be detected by typical cryptographic mechanisms \cite{van2018survey}. However, the current state of the art for misbehavior detection is usually based on supervised learning techniques \cite{uprety2021privacy,boualouache2022federated,boualouache2022survey} that need labelled datasets to train a model. This process might require human intervention becoming a very resource-intensive and time-consuming activity to have numerous labelled examples in order to achieve a proper generalization \cite{karoly2018unsupervised, saravanan2018state}. In the case of vehicular environments, the creation of such labelled dataset could be infeasible due to the impossibility to reproduce a real scenario. Furthermore, supervised learning techniques are not appropriate to detect zero-day attacks, which refer to previously unknown threats. Instead, the use of unsupervised learning techniques could be used to mitigate such concerns \cite{usama2019unsupervised, zhang2023federated}. In particular, unsupervised learning is used to extract information on the data’s structure and hierarchy by using the data samples without the need for ground truth files. The extracted knowledge representation can be used as a basis for a deep model that requires fewer labelled examples \cite{karoly2018unsupervised, jin2023federated}. 

In spite of the advantages provided by unsupervised learning, we notice that most of the proposed approaches for misbehavior detection are based on supervised techniques using unrealistic data distributions considering an FL setting \cite{boualouache2022survey, hernandez2023intrusion}. Indeed, some of the recent works \cite{uprety2021privacy} are based on artificial divisions of vehicular misbehavior datasets, such as the Vehicular Reference Misbehavior (VeReMi) dataset \cite{van2018veremi}. To fill this gap, our work proposes a combination of clustering techniques and autoencoders (AE) for misbehaviour detection.  On the one hand, clustering algorithms use unlabelled data to create clusters that achieve high inner similarity and outer dissimilarity, without relying on signatures or explicit descriptions of attack classes. In particular, we choose Gaussian mixture models (GMM) as clustering technique since it adds probabilities, therefore, removing the restriction that one point has to belong only to one cluster. Additionally, GMM-based clustering also performs better in several scenarios where other clustering methods do not provide suitable results \cite{vashishth2019gmmr}. On the other hand, we use a specific type of AE called Variational Autoencoder (VAE) \cite{bishop1994neural}. The main goal of an AE is to reconstruct the input data and the variational version is based on a stochastic encoder to map the input to a probability distribution. Since they achieve great success in generating abstract features of high-dimensional data, the detection of abnormal samples increases significantly \cite{said2020network}. Additionally, we implement Restricted Boltzmann Machines (RBM), which represent a type of neural network (NN) used as pre-training layer for the VAE to foster convergence during the federated training process. To the best of our knowledge, this is the first work exploring the use of such techniques in the scope of vehicular misbehavior detection. Furthermore, we use the VeReMi dataset for evaluation purposes by considering a realistic FL setting where each vehicle is intended to train with its local data. In our system, the server (or aggregator) is hosted in the cloud, which will serve as a centralized hub for refining and updating misbehavior detection models by the aggregation of the data process carried out by the vehicles acting as clients. By using a cloud-based approach, the system can easily scale to accommodate a growing fleet of vehicles, making it a practical solution for large deployments. Indeed, it is intended to store a track record of detected misbehaviour attacks, and to provide real-time monitoring and analysis of model performance across the entire set of clients. We carry out an exhaustive comparison between regular AE and VAE, and unlike most existing FL-enabled misbehavior detection approaches, we use Fed+ \cite{yu2020fed+} as the aggregation function to deal with non-iid scenarios \cite{campos2021evaluating}. Our contributions are summarised as follows:

\begin{itemize}
    \item Unsupervised FL approach based on VAE and GMM considering a realistic partition of the VeReMi dataset for misbehavior detection.
    \item Use of RBMs to enhance the convergence of VAEs in the FL setting.
    \item Application of Fed+ as aggregation function to improve the effectiveness of the approach in the presence of non-iid data distributions.
    \item Comprehensive evaluation to demonstrate the effectiveness of the proposed approach and comparison with recent works using the VeReMi dataset.
\end{itemize}

The structure of the paper is as follows: Section \ref{sec:Preeliminaries} describes the main concepts and techniques used in our work. Section \ref{sec:Related} analyses other works related to the application of GMM and AEs, as well as the use of AEs considering FL settings. Then, in \ref{sec:System} we provide a detailed description of our misbehavior detection system. Section \ref{sec:Evaluation} describes the results of our proposed approach and compares it with recent works using the same dataset. Finally, Section \ref{sec:Conlutions} concludes the paper.

\section{Preliminaries}\label{sec:Preeliminaries}
In this section, we describe the main operation of FL, as well as the ML models we used in our approach, including GMM, and VAE.

\subsection{Federated learning (FL)}
Traditional ML approaches are characterized by centralized deployments where different parties are forced to share their data to be analyzed. Such a scenario poses significant issues around the needs of network connectivity, latency, as well as compliance with existing data protection regulations. To overcome such problems, FL \cite{mcmahan2017communication} was proposed as a decentralised approach to train a ML model ensuring that data are not shared. Indeed, Federated learning (FL) scenarios are usually represented by two main components: the central entity (or server), and the data owners (or clients). Communication between these parts is essential for the correct creation of the model. The key of FL is that clients are intended to create a model in a collaborative way by only training with their local data and sharing the model updates through the server. 

 \begin{figure} [!ht]
	\centering
		\includegraphics[width=\columnwidth]{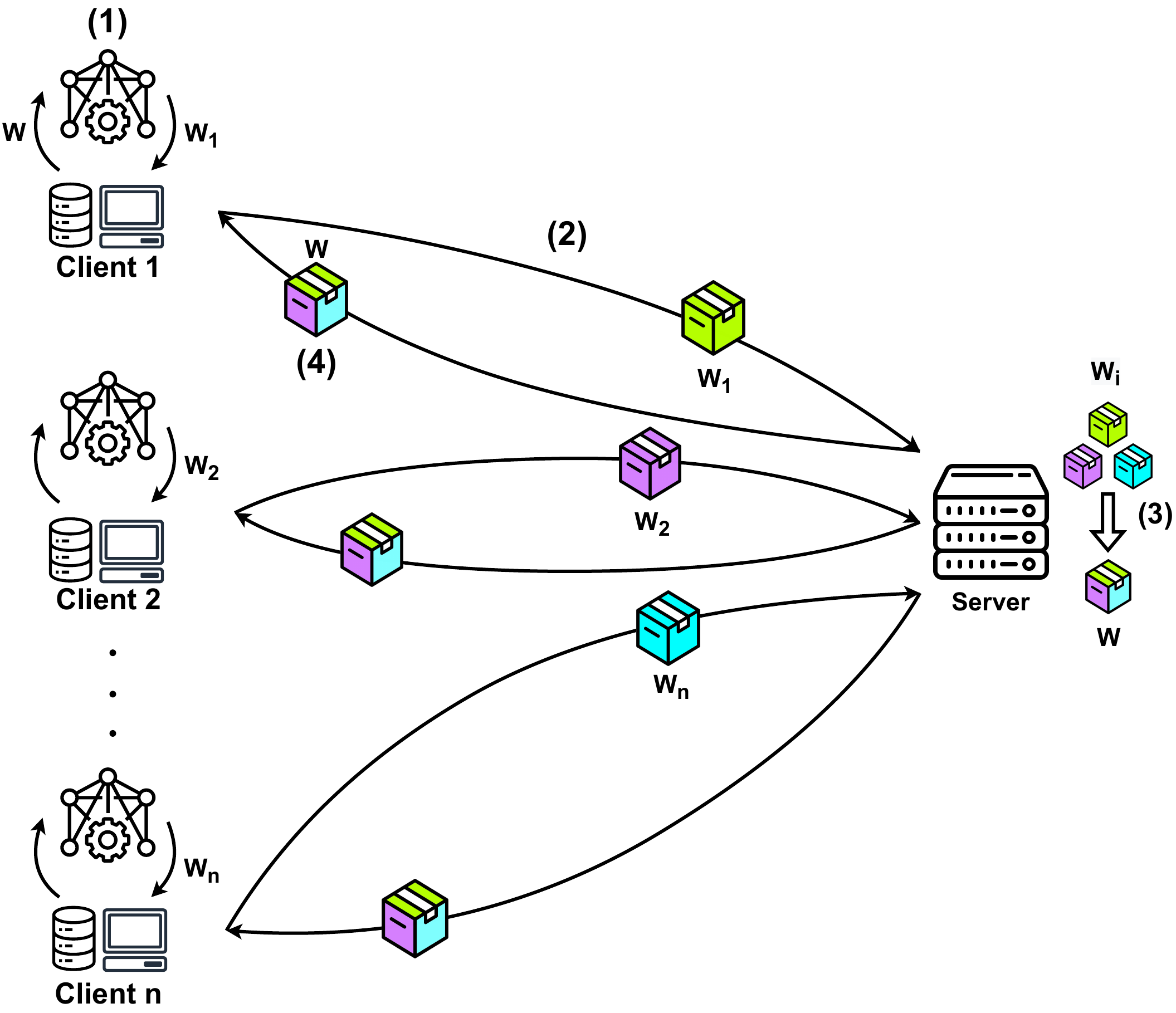}
	\caption{Pictorial description of an FL setting}
	\label{fig:FLscheme}
\end{figure}

A visual description of the FL operation is displayed in Fig. \ref{fig:FLscheme} where (1) clients train the model using their dataset. Next, (2) they send the weights resulting weights from the local training to the server, which uses an aggregation function (3) to update the clients' incoming weights. Finally, (4) the server sends the aggregated weights to the clients, which starts again another training round through (1). For aggregating the weights in the server, our approach is not based on the well-known FedAvg function \cite{mcmahan2017communication}, but on the Fed+ \cite{yu2020fed+}, which is able to provide better performance in non-iid scenarios compared to FedAvg \cite{campos2021evaluating}. The main characteristic of Fed+ is that parties are not forced to converge to a single central point, as explained below. Hence, the weights are uploaded following the following equation:
 \begin{equation} \label{eq:Fed+weights}
    W^{k+1} = \theta[W^k - \nu \nabla f_k(W^k)] + (1-\theta)Z^k,
\end{equation}
where $\nu$ is the learning rate, $\theta$ is a constant between 0 and 1 that controls the degree of regularisation, and $Z^k$ is the average of clients' weights. As the new weights are an average of local weights and global weights, this means that each party has different weights and does not depend on the fact that all clients converge to the same point.

\subsection{Gaussian mixture models}
A Gaussian mixture model (GMM) is a probabilistic model that assumes all the data points are generated from a mixture of a finite number of Gaussian distributions with unknown parameters. Formally, a GMM of $K$ components is a parametric probability density function represented as a weighted sum of $K$ normal distributions, i.e.:
$$p(x|\mu_1,\dots,\mu_K,\sigma_1,\dots,\sigma_K,\pi_1,\dots,\pi_K)=\sum_K \pi_i \mathcal{N}(\mu_i,\sigma_i)$$

where $\mu_i$ are the means, $\sigma_i$ the variances, $\pi_i$ the proportions weights (which sum one and are positives), and $\mathcal{N}$ is a  Gaussian with specified mean and variance. A Gaussian distribution is completely determined by its covariance matrix and its mean. The covariance matrix of a Gaussian distribution determines the directions and lengths of the axes of its density contours, all of which are ellipsoids. There are 4 types of covariance matrix: full, tied, diagonal, and spherical. Full means the components may independently adopt any position and shape, tied means they have the same shape, but the shape may be anything, diagonal means the contour axes are oriented along the coordinate axes, but otherwise the eccentricities may vary between components, and spherical is a "diagonal" situation with circular contours. The purpose of using GMMs is the division of the dataset observations into $K$ groups or clusters. Due to their probabilistic nature, GMMs differ from other clustering methods such as K-means \cite{awasthi2012improved} in the fact that an observation can belong to more than one cluster. More specifically, GMMs estimate the likelihood of every observation belonging to each component and assign the observation to the most likely one. In our work, we exploit these probabilities to classify samples as anomalous or not anomalous. Additionally, GMMs present other advantages such as adapting to the shape of the clusters, handling missing data, and taking into account the variance of the data. These properties are described in depth in  \cite{reynolds2009gaussian}.

\begin{figure} [!ht]
	\centering
		\includegraphics[width=\columnwidth]{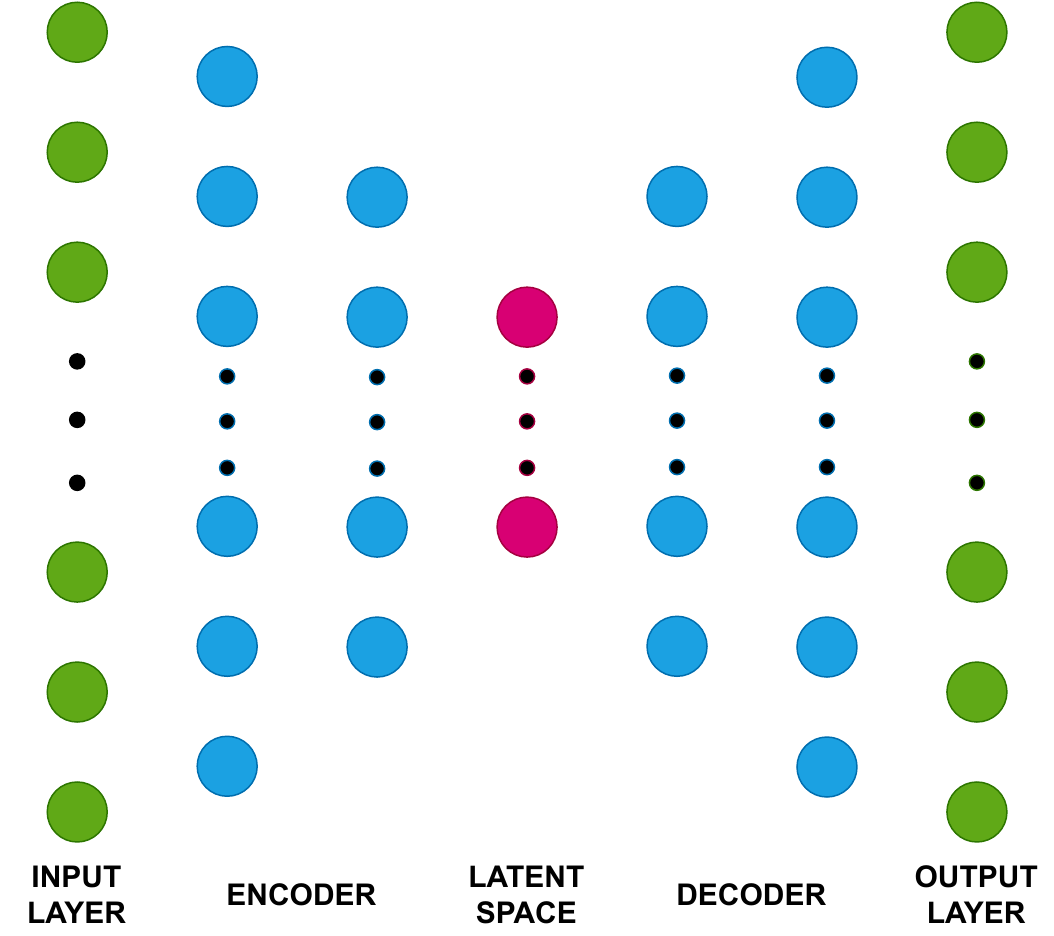}
	\caption{AE's pictorial description. It consists of a symmetric fully connected NN where the input is compressed through the encoder to the latent space to be reconstructed by the decoder.}
	\label{fig:autoencoderImage}
\end{figure}

\subsection{Variational autoencoder (VAE)}
An autoencoder (AE) \cite{bishop1994neural} is a particular case of NN where the input and output dimensions coincide, and its layer structure is symmetric. AEs are used to build models by using unlabelled data; therefore, they are an example of unsupervised learning techniques. An AE consists of an encoder and a decoder, as shown in Fig. \ref{fig:autoencoderImage}. The encoder takes the input data and compresses it through the hidden layers into a lower dimension code, called \textit{latent space} ($Z$). Then, the decoder reconstructs this latent space into its original state, that is, the input data. The main goal of AEs is to create a copy as close as possible to the original data. The difference between the input and output is called reconstruction error, which is usually measured using the Root Mean Square Error (RMSE). In a scenario for misbehavior or attack detection, training the model with only benign data will produce a high reconstruction error when attack data is passed. Hence, a threshold is set to decide which samples are benign or could be considered as an attack. 

A Variational AE (VAE) \cite{kingma2013auto} is a type of AE whose encoding distribution is regularized, i.e., it approaches a standard normal (or gaussian)  distribution during the training to ensure that its latent space is able to generate new data \cite{zhang2021federated, yao2019unsupervised, sun2018learning, an2015variational}. The main difference between a VAE and an AE is that the VAE is a stochastic generative model that provides calibrated probabilities, while a simple AE is a deterministic discriminative model without a probabilistic foundation. AEs are not suitable to generate new samples so it can cause the decoder to provide misleading results since all the information condensed in the latent space is disorganised. For that reason, the VAE is intended to map the encoder distribution $q(Z|X)$ to a standard normal distribution for providing more order in the latent space. For this purpose, the \textit{reparameterization trick} is carried out. From this distribution $q(Z|X)$, it takes its mean $Z_\mu$ and standard deviation $Z_\sigma$. With these two latent variables, a sample is created to the latent variable $Z$ that is sent to the decoder in order to create the predicted output $\hat{X}$. This method also ensures that the network can be trained using backpropagation.  

\begin{figure} [!ht]
	\centering
		\includegraphics[width=\columnwidth]{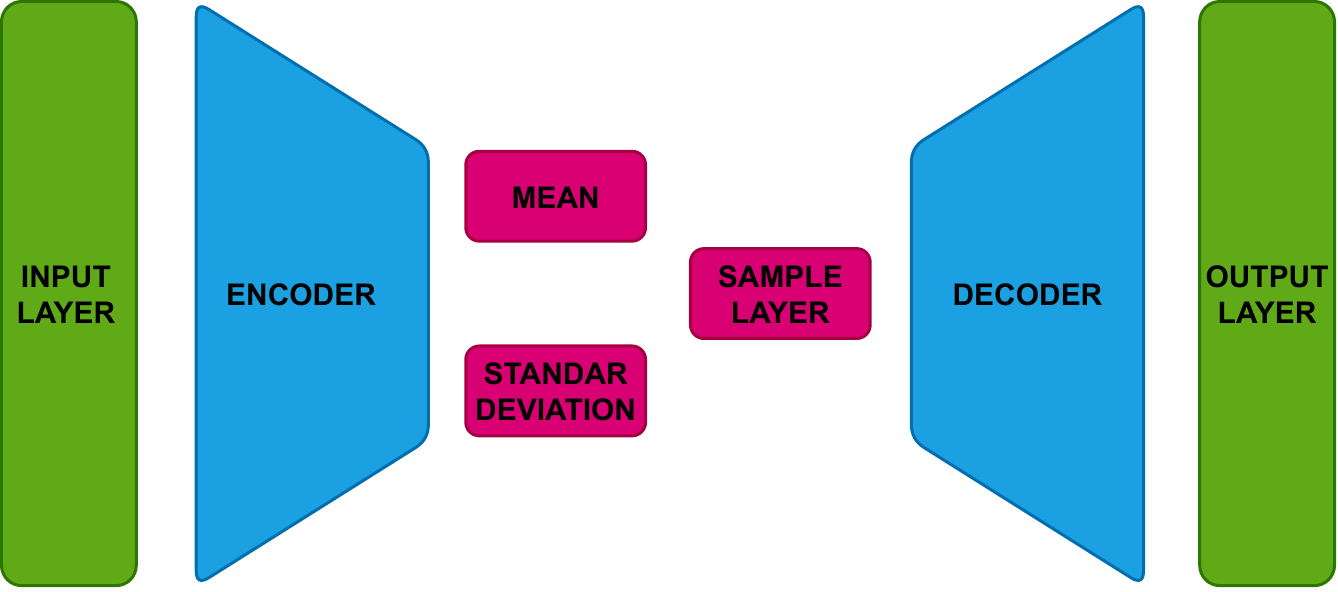}
	\caption{Pictorial description of a VAE. Similar to the AE, but the latent space is changed by three layers to ensure the encoder follows a normal distribution.}
	\label{fig:VAEimage}
\end{figure}

Once the VAE has been described, the loss function has to be set.  The main goal of VAE is to create a copy of an input vector. Hence, the RMSE has to be minimized. At the same time, as said previously, the model distribution needs to be as close as possible to a standard gaussian. For this, the Kullback–Leibler Divergence \cite{joyce2011kullback} (KL-divergence) function is used. This function measures how similar two distributions are. Hence, this difference between the encoder distribution and a standard distribution  using the KL-divergence has to be as close as 0 as possible. Therefore, the loss function of VAE is:

\begin{equation}
    Loss(X,\hat{X}) = Loss_{RMSE}(X,\hat{X}) + KL(q(Z|X),\mathcal{N}(0,1))
\end{equation}

One of the challenges associated with training AEs is the issue of convergence, especially when starting from different initial states. This problem becomes even more visible in a Federated Learning (FL) setting, where non-iid data distributions are prevalent among the participating devices or clients. To alleviate the convergence issues, researchers have turned to Restricted Boltzmann Machines (RBMs) as a promising solution. RBMs have gained recognition in the literature for their role in initializing AEs effectively, as proposed by Hinton and Salakhutdinov in their work \cite{hinton2006reducing}. RBMs are a type of stochastic neural network with a distinctive two-layer architecture. They are characterized by symmetric connections between the layers and the absence of self-feedback loops. Importantly, RBMs exhibit full connectivity between the two layers while having no connections within a layer.

The two layers in an RBM are referred to as the visible layer and the hidden layer. The visible layer contains the original input data, while the hidden layer captures higher-level representations and features. This architecture makes RBMs well-suited for various applications, including feature extraction, pattern recognition, dimensionality reduction, and data classification.

The energy function of an RBM is defined as:

\[E(\mathbf{v}, \mathbf{h}) = -\sum_{i=1}^{N_v} \sum_{j=1}^{N_h} w_{ij} v_i h_j - \sum_{i=1}^{N_v} a_i v_i - \sum_{j=1}^{N_h} b_j h_j\]

Where:
\begin{itemize}
    \item \(\mathbf{v}\) is the visible layer's binary state vector with \(N_v\) neurons.
    \item \(\mathbf{h}\) is the hidden layer's binary state vector with \(N_h\) neurons.
    \item \(w_{ij}\) represents the weight connecting visible neuron \(i\) to hidden neuron \(j\).
    \item \(a_i\) is the bias of the visible neuron \(i\).
    \item \(b_j\) is the bias of the hidden neuron \(j\).
\end{itemize}

The joint probability of a configuration (\(\mathbf{v}\), \(\mathbf{h}\)) in an RBM is given by the Boltzmann distribution $P(\mathbf{v}, \mathbf{h}) = \frac{e^{-E(\mathbf{v}, \mathbf{h})}}{Z}$.

Where:
- \(Z\) is the normalization constant (partition function) that ensures the probabilities sum up to 1 over all possible configurations.

The conditional probability of the hidden layer given the visible layer is defined as $P(\mathbf{h} | \mathbf{v}) = \prod_{j=1}^{N_h} P(h_j | \mathbf{v})$. And similarly, the conditional probability of the visible layer given the hidden layer is $P(\mathbf{v} | \mathbf{h}) = \prod_{i=1}^{N_v} P(v_i | \mathbf{h})$.

Where the conditional probabilities for binary units are typically sigmoid functions:

\[P(h_j = 1 | \mathbf{v}) = \sigma\left(\sum_{i=1}^{N_v} w_{ij} v_i + b_j\right)\]

\[P(v_i = 1 | \mathbf{h}) = \sigma\left(\sum_{j=1}^{N_h} w_{ij} h_j + a_i\right)\]

Here, \(\sigma(x)\) is the sigmoid activation function:

\[\sigma(x) = \frac{1}{1 + e^{-x}}\]

In this context, we adopt a similar strategy as outlined in previous studies \cite{wu2021research, pacheco2018restricted, saravanan2020deep}. Specifically, we employ RBMs for pre-training Variational Autoencoders (VAEs). This pre-training step helps the VAE initialize its parameters in a way that is more likely to converge effectively during subsequent fine-tuning. By leveraging RBMs for this purpose, we aim to enhance the overall robustness and convergence behavior of AEs within the FL paradigm, ultimately contributing to the successful deployment of federated machine learning systems in non-iid data environments.

\section{Related Work}\label{sec:Related}

The use of ML techniques for misbehavior detection has attracted significant attention in recent years \cite{boualouache2022survey}. However, as previously described, most of the proposed works rely on centralized approaches using supervised learning techniques. In this direction, \cite{sharma2020machine} uses six different supervised techniques along with plausibility checks to come up with a multiclass classification approach on the VeReMi dataset. Using the same dataset, \cite{slama2022comparative} also discusses the use of different supervised techniques considering various feature extraction methods. Additionally, \cite{anyanwu2021real} uses an optimized version of Random Forest (RF), which is compared with other techniques such as K-Nearest Neighbors (KNN) and Decision Trees (DT). 

As an alternative to centralized ML approaches, recently the use of FL has been also considered in the scope of misbehavior detection for vehicular environments. Indeed, \cite{uprety2021privacy} proposes a federated approach based on neural networks to build a multiclass classification approach for detecting the attacks contained in the VeReMi dataset. Furthermore, \cite{moulahi2022privacy} proposes a federated approach using blockchain where several supervised learning techniques are tested on an extended version of such dataset \cite{kamel2020veremi}. In the case of \cite{kristianto2023misbehavior}, the authors propose an approach based on a semi-supervised model using a neural network and a subset of labeled data for the initial training phase. Then, new unlabeled data is used to improve the effectiveness of the system. The approach is also applied to the extension of the Veremi dataset.

Unlike previous approaches, our work offers an unsupervised approach for detecting misbehavior using a model based on the combination of VAEs and GMM. Although this approach has been scarcely considered in this field, it has been widely used for detecting different types of attacks and anomalies. In this direction, a Deep Autoencoding Gaussian Mixture Model (DAGMM) for unsupervised anomaly detection is proposed by \cite{zong2018deep}. DAGMM consists of a deep AE to reduce the dimensionality of input sample, and a GMM that is fed with the low-dimensional data provided by the AE. Furthermore, \cite{an2022ensemble} describes an approach based on AE and GMM in which the objective functions’ optimization problem is transformed into a Lagrangian dual problem. Indeed, authors make use of the Expectation-Maximization (EM) algorithm. Like in the previous work, AEs are used for input’ dimensionality reduction. Then, an estimation network is intended to estimate samples’ density, so that samples with density higher than a certain threshold are identified as outliers. Moreover, the use of VAEs and GMM is considered by \cite{liao2018unified}. The VAE first trains a generative distribution and extracts reconstruction-based features. Then, the authors use a deep belief network to estimate the mixture probabilities for each GMM’s component. Based on these results, the GMM is used to estimate sample densities with the EM algorithm.

While previous works are based on centralized settings, other recent approaches were proposed considering the use of VAEs/AEs and GMM on a FL scenario. Similar to the approach made by \cite{liao2018unified}, however, in the context of encrypted aligned data, the recent work \cite{zhao2024deep} use VAE for reducing the dimension of the data, and then GMM is applied for clustering the data and selecting the correct samples. As we will see in section \ref{sec:Evaluation}, the application on GMM without implementing VAE during the classification will lead to misclassification to a certain types of samples. Looking at works using AEs, a federated version of the previously described approach DAGMM is proposed by \cite{chen2019network}, so that each client is intended to share the AE’s updated weights in each training round. Other works, such as \cite{preuveneers2018chained, cetin2019federated, alazzam2022federated} also use AEs in a federated scenario considering different contexts, such as wireless sensor networks. Like in the previous case, anomaly detection is based on the reconstruction error, so that samples with higher values are considered anomalies. Moreover, \cite{wu2022fl} proposes MGVN that represents an anomaly detection classification model using FL and mixed Gaussian variational self-encoder, which is built by using a convolutional neuronal network. 

In general, our analysis of existing literature reflects a lack of approaches applying unsupervised learning techniques for misbehavior detection and other cybersecurity-related problems.  Furthermore, the described works are based on similar approaches in which AEs are used initially and the resulting output feeds the GMM. In our case, we initially train the GMM with the input data, and the results are used by the VAE to carry out a federated training among the vehicles acting as FL clients. To the best of our knowledge, this is the first work considering the use of GMM and VAEs in the contest of vehicular misbehavior detection. Furthermore, unlike existing works applying ML techniques in this context, our approach considers a non-iid data distribution scenario in which each vehicle trains on its local data. To mitigate the impact on non-iid in the approach’s performance, we further apply an approach to balance the dataset and use Fed+ as the aggregation function as described in the following sections.

\section{Proposed misbehavior detection system}\label{sec:System}
This section provides a detailed description about our unsupervised FL-enabled misbehavior detection approach. Furthermore, we describe the dataset used and how it is preprocessed below. 

\subsection{Dataset and preprocessing} \label{dataandprep}
For evaluation purposes, we use the VeReMi \cite{van2018veremi} and VeReMi extension \cite{kamel2020veremi} datasets. On one hand, VeReMi is a simulated dataset that was created using VEINS and the Luxembourg SUMO Traffic (LuST) scenario \cite{7385539}. On the other hand, VeReMi extension make use of Framework For Misbehavior Detection (F²MD). F²MD is a VEINS extension that enables the recreation and detection of various misbehavior detection use cases. LuST is a traffic simulation scenario validated with real driving data. VeReMi utilizes a subsection of the LuST scenario. VeReMi dataset was created by simulating 225 scenarios considering 5 position forging attacks, 3 vehicle densities (low, medium and high), and 3 attacker densities (10, 20 and 30 percent), and each parameter set was repeated 5 times for randomization. Finally, the dataset contains the log messages of 498 vehicles. For VeReMi extension, it uses subsection of the LuST
network with a size 1.61 \textit{km²} and a peak density of 67.4 \textit{Veh/km²}. Furthermore, VeReMi extension contain two vehicle densities, one in the rush hour time (7h-9h), and another in low traffic time (14h-16h). Each file maintains a record of Basic Safety Messages (BSMs) received by a single vehicle (same ID) from neighbouring vehicles (300-meter range) during its entire journey. The files are converted to CSV format differentiating between benign and attack samples. 

Furthermore, we carry out a pre-processing of the VeReMi dataset to deal with the impact of non-iid data distributions on the effectiveness of the misbehavior detection approach. The main reason is that our approach (unlike other recent works, such as \cite{moulahi2022privacy}) considers each vehicle as an FL client using its own data during the federated training process. Indeed, as shown in our previous works \cite{campos2021evaluating}, non-iid data distributions might have a dramatic impact on the effectiveness of the FL-enabled system. For this purpose, we use SMOTE-Tomek \cite{batista2003balancing}, which represents a combination of oversampling and undersampling techniques to balance the class ratio. The new samples in the dataset are created through SMOTE by making linear combinations between close points of the less-represented class. Then, Tomek Links is in charge of removing neighbour points that belong to different classes in order to make the classes more differentiable. Initially, the balance of the clients' dataset is 75\% benign-25\% attack. After applying SMOTE-Tomek, the balance is nearly 50-50 in all clients. Finally, as explained in section \ref{sec:Preeliminaries}, GMM are a sum of normal distributions. Therefore, the dataset has to follow a Gaussian distribution. In this sense, we normalise the dataset and check whether it fits to a Gaussian distribution by applying the Shapiro-wilk test \cite{SHAPIRO1965}. After applying the Shapiro-wilk test we have  a Shapiro-wilk score of 0.98 and $p$-value of $0.99>0.05$, meaning that the dataset follows a Gaussian distribution.

\subsection{System description}
After describing the dataset and the pre-processing we carry out, below we provide a detailed explanation of the proposed system, including the different phases of the misbehavior detection approach. 

\subsubsection{Overview}
\begin{figure*} [!ht]
	\centering
		\includegraphics[width=2\columnwidth]{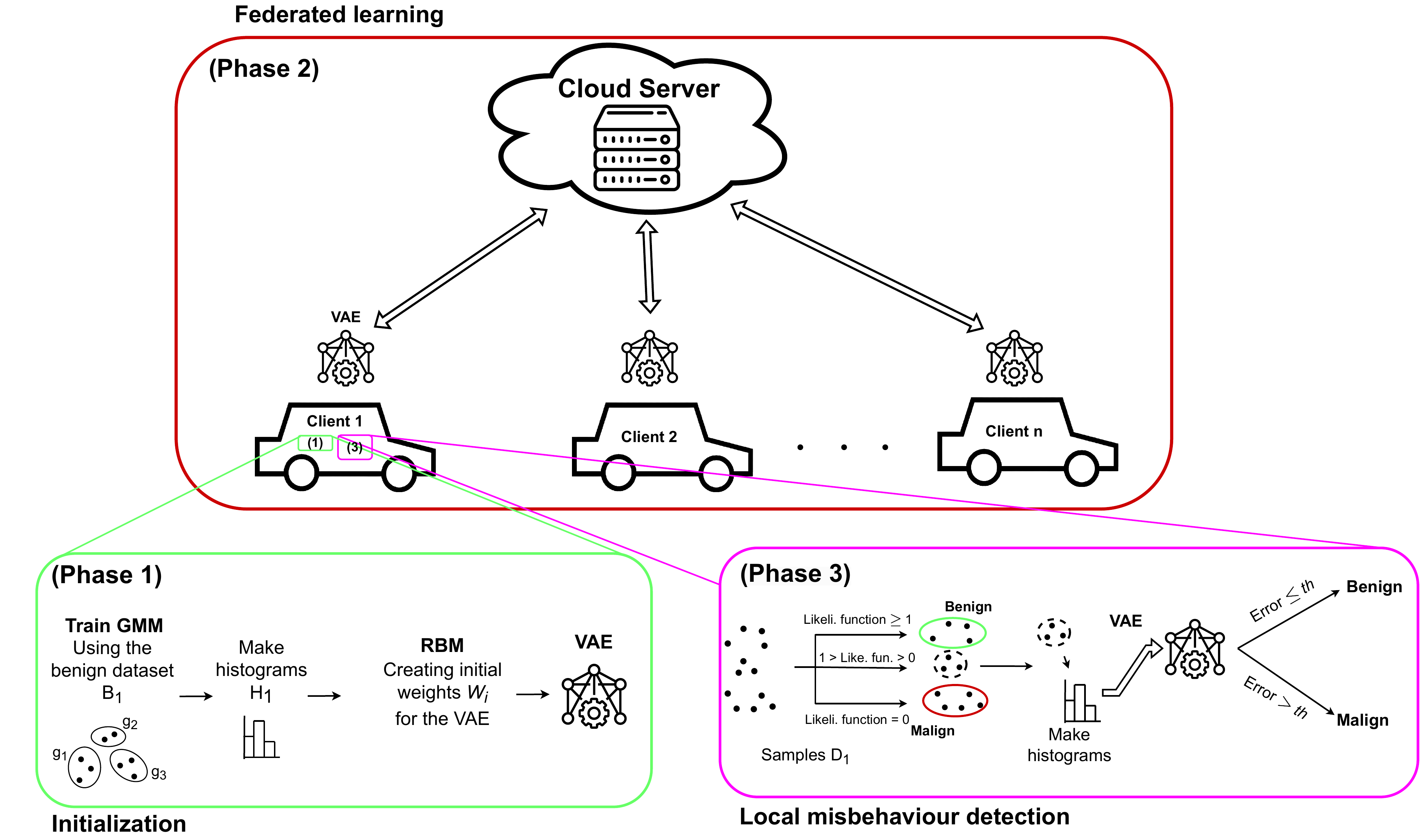}
	\caption{Pictorial description of our misbehavior detection system. There are three phases, the first one (1), is the initialization, where the clients train the GMM and create the histograms subsequently. Also, it uses RBM to create VAE's initial weights. Then, in (2), clients create a federated environment to train their VAE models. And finally, in (3), the local misbehaviour detection is carried out, where each client classifies the samples as benign or malign using the model trained and the threshold $th$ set.}
	\label{fig:systemModel}
\end{figure*}

A global overview of our system is shown in Fig. \ref{fig:systemModel}, which includes the relationship between the different components and techniques previously described. As already mentioned, vehicles act as FL clients performing local training by using the VeReMi dataset and the global model that is updated in each training round. In particular, our approach consists of 3 phases. Firstly, during Phase 1 (Initialization), clients train the GMM with their benign dataset. Then, this data is transformed into histograms that are used to create the initial weights for training the VAE by using the RBMs. For each sample, these histograms measure which features are within each group created by the GMM. Once each client has initialized its corresponding VAE, Phase 2 (Federated learning) is carried out by following the steps described in Section \ref{sec:Preeliminaries}, so that each client performs a local training by using its own data in each training round. The server implements an aggregation method to aggregate the local results from the FL clients, and the resulting aggregated model is sent again to the FL clients to be updated through a new training round. While the vehicles themselves act as FL clients, the server is intended to be deployed on the cloud. Therefore, FL is employed to detect cyberattacks locally in these vehicles, and the server in the cloud  can then aggregate data from various vehicles and further analyze it to gain a holistic view of the cybersecurity landscape. It should be noted that this cloud service could be also considered to store information about vehicles that were identified as misbehaving entities by our detection system, so that it allows the system to keep track of potential malicious vehicles. In Phase 3 (Local misbehavior detection), after the federated training process is finished, each client has a trained VAE to be used for classification purposes in order to detect potential misbehavior. In particular, the GMM's likelihood function \cite{etz2018introduction} is used to evaluate whether a point belongs or not to one of the groups previously created in Phase 1. Depending on the likelihood value, the sample is considered a benign sample (greater than 1) or an attack (0). In case of a value between 0 and 1, a histogram is created for that sample, and the local VAE is used to classify it depending on a certain threshold value. These processes are further detailed in the following subsections.

\subsubsection{Model training}
As previously mentioned, the model training process is split into two main phases: Initialization and Federated learning. Before starting this process, it should be noted that, for each client, we pre-process its corresponding dataset as described in subsection \ref{dataandprep}, and we use the 80\% of the benign dataset to train the GMM with the number of components previously optimised. Toward this end, we use the Silhouette coefficient \cite{rousseeuw1987silhouettes}, which represents a metric to calculate the performance of a particular clustering by examining how separated and compact the clusters are. This optimization will lead to each client will have different numbers of components, as it will be explain in section \ref{sec:Evaluation}, due to this number of components define the structure of the FL model, the clients are divided in groups regarding their number of components. In this direction, although there will be several gruops, the model training scheme will be the same in each case. 

The model training process is further detailed in Algorithm \ref{alg:train}. Firstly, in Phase 1, for each client $n$, GMM is used to divide the benign dataset $B_n$ as many groups $G_n=(g_n^k)_1^{K_n}$ as components $K_n$ are previously computed (line 2). Then, the data is transformed into histogram vectors  \cite{zolotukhin2016increasing} (line 3), which are used in our previous works \cite{garcia2021distributed}. This technique creates a matrix $H_n$ containing the vectors $h^i = (h^i_1,h^i_2,\dots,h^i_{K_n})$ (line 21), where $K_n$ is the number of groups (equal to the number of GMM components), and each $h^i_k$ is the number of features of sample $i$ that are within each group $g^k_n$ divided by the total number of features (line 15). To calculate it, for each feature $j$ of sample $i$ we check whether the value of this feature is within the dimension $j$ of the cluster $g_n^k$ (lines 12-14), i.e, whether is between the dimension $j$ of the center of this cluster $g_n^k$ minus the standard deviation and the dimension $j$ of the center of the cluster $g_n^k$ plus the standard deviation. The standard deviation means the standard deviation of the points of $g_n^k$ Furthermore, it should be noted that function $Dim_j(x)$ is only intended to take the coordinate $j$ of point $x$. The resulting histogram $H_n$ is used by the RBM to pre-train the VAE in order to get the initial weights $W^0_n$ (lines 4-5) for the next steps. 

In Phase 2, the FL process is carried out by the different clients which have the same number of components using the local VAE models (line 7), whose input dimension is the number of components previously set. Each client is intended to train locally its corresponding VAE model and share the resulting weights in each training round (lines 22-32) with the server, as previously described in Section \ref{sec:Preeliminaries}. Once this process is complete after a certain number of rounds, we set the threshold as:

\begin{equation}\label{eq:RecErr}
    th = mean(RE)+0.01*std(RE)
\end{equation}
where $mean(RE)$ and $std(RE)$ are the mean and standard deviation of the VAE's reconstruction error (RE) respectively (line 10). We use this formula since it provides a value close to the 95-percentile.

\begin{algorithm}
\caption{Model training}\label{alg:train}
\begin{algorithmic}[1]
\REQUIRE{$N$ set of clients with same number of components $K_n$, $B_n$ benign dataset of client $n$, $GMM_n$ GMM of client $n$, $VAE_n$ VAE model of client $n$, $rounds$ number of rounds in the FL environment, $threshold$ is the eq. \ref{eq:RecErr}, $Fedplus$ is the eq. \ref{eq:Fed+weights}, and $\theta$ the Fed+ constant.}
\ENSURE{VAE model trained and threshold of each client}\hfill \break
\textbf{(Phase 1)}
\FOR{$n \in N$}
\STATE $G_n = GMM_n(B_n) = (g_n^k)_1^{K_n}$
\STATE $H_n =$ Create histograms($B_n$, $G_n$)
\STATE $ W^0_n = RBM_n(H_n)$
\STATE Set $W^0_n$ to $VAE_n$
\ENDFOR

\vspace{5pt}
\textbf{(Phase 2)}
\STATE $FL((VAE_n)_1^N,(H_n)_1^N, rounds)$
\STATE $th_n = threshold(VAE_n) \forall n \in N$ 

\vspace{5pt}
Create histograms($B_n$, $G_n$):
\FOR{ $i \in B_n$}
\FOR{$g\in G_n$}
\FOR{$j \in features(B_n$)}
\STATE Sup\_bound = $Dim_j(Center_g) + Dim_j(Std_g)$
\STATE Inf\_bound = $Dim_j(Center_g) - Dim_j(Std_g)$
\IF{$j \in [Inf\_bound, Sup\_bound]$}
\STATE $h_g^i += \frac{1}{length(features)}$
\ENDIF
\ENDFOR
\ENDFOR
\STATE $h^i=(h^i_1,\dots, h^i_{length(G_n)})$
\ENDFOR
\STATE \textbf{return: }$H=(h^i)_1^{length(B_n)}$ \hfill \break

FL($(VAE_n)_1^N,(H_n)_1^N, rounds$):
\FOR{$r$ in [1,$\dots$, $rounds$]}
\IF{$r>1$}
\STATE Server sends weights $Z$ to clients, otherwise the clients use $W^0_n$
\ENDIF
\FOR{$n \in N$}
\STATE Train $VAE_n$ with $H_n$ 
\STATE $\text{Get weights }W^r_n\text{ of }VAE_n$
\STATE $W_n^{r+1} = FedPlus(Z, W_n^r, \theta)$
\STATE $Client_n\text{ sends }W_n^{r+1}\text{ to server }$
\ENDFOR
\STATE \text{Server aggregates all the }$(W_n^{r+1})_1^N$\text{ into }$Z$.
\ENDFOR
\STATE \textbf{return: } $(VAE_n)_1^N$ trained
\end{algorithmic}
\end{algorithm}

\subsubsection{Local misbehavior detection}
After the previous steps are complete, each vehicle is equipped with a model to detect potential misbehavior detection (Phase 3 in Figure \ref{fig:systemModel}) based on the classification of samples during the testing process of the model. For this purpose, each vehicle employs the portion of benign data that was not used for training (i.e., 20\%), as well as the same size of attack data. The process is detailed in Algorithm \ref{alg:cap}, which is based on the calculation of the likelihood function associated with the GMM  for each sample of a vehicle's testing dataset (line 2). If this value is greater than or equal to 1 (line 3), it means that such sample belongs to one of the groups created by the GMM, so it is classified as benign (line 4). If it is equal to 0 (line 5), it does not belong to a GMM group, so it is classified as anomalous (line 6). Then, the samples whose value is between 0 and 1 (line 7) could represent attacks mirroring normal traffic \cite{zolotukhin2016increasing}. These samples are transformed into the histograms (line 8) as previously described. Then, the VAE that was trained in the previous phase is applied to reconstruct the histogram (line 9); if the reconstruction error is higher than the threshold (line 11), it means that it is anomalous (line 12), otherwise, such sample is considered as benign data (line 14).

\begin{algorithm}
\caption{Local misbehavior detection (Phase 3)}\label{alg:cap}

\begin{algorithmic}[1]
\REQUIRE{$D_n$ Testing dataset of client $n$, $LF_{GMM_n}$ likelihood function of  gaussian mixture model of client $n$ $GMM_n$, $G_n$ groups created by $GMM_n$, $VAE_n$ VAE model of client $n$, $RMSE$ is the Root Mean Square Error, $th_n$ threshold set by the client $n$\hfill \break}
\ENSURE{  Accuracy of the model $a$\hfill \break} \\
\FOR{$x\in D_n$}
\STATE $p = LF_{GMM_n}(x)$
\IF{$p\geq 1$}
\STATE $x$ is benign
\ELSIF{$p= 0$}
\STATE $x$ is anomalous
\ELSE
\STATE  $H=Histogram(x, G_n)$
\STATE $\hat{H}= VAE_n(H)$
\STATE $RE_n(\Hat{H})= RMSE(H, \hat{H})$ 
\IF{$RE_n(\hat{H}) > th_n$}
\STATE $x$ is anomalous
\ELSE
\STATE $x$ is benign
\ENDIF
\ENDIF
\ENDFOR 
\STATE Calculate confusion matrix $M$ and get accuracy.\hfill \break

$Histogram(x, G_n):$
\FOR{$g\in G_n$}
\FOR{$j \in features(x$)}
\STATE Sup\_bound = $Dim_j(Center_g) + Dim_j(Std_g)$
\STATE Inf\_bound = $Dim_j(Center_g) - Dim_j(Std_g)$ 
\IF{ $j \in [Inf\_bound, Sup\_bound]$}
\STATE $h_g^i += \frac{1}{length(features)}$
\ENDIF
\ENDFOR
\ENDFOR
\STATE \textbf{return: }$h^i=(h^i_1,\dots, h^i_{length(G_n)})$
\end{algorithmic}
\end{algorithm}

\section{Evaluation}\label{sec:Evaluation}
For the evaluation of our misbehaviour detection approach, we employ a virtual machine with processor Intel(R) Xeon(R) Silver 4214R CPU @ 2.40GHz with 24 cores and 196 GB of RAM. Furthermore, we use Flower \cite{flower} as the FL implementation framework, the version 0.18. Flower provides an end-to-end implementation and  the possibility to customise several aspects of the model for performing a more exhaustive evaluation. All the parameters related to our federated scenarios are summarised in Table \ref{tab:summary_fl}. Furthermore, it should be noted that we make the code and the dataset used publicly available for reproducibility purposes\footnote{\url{https://github.com/Enrique-Marmol/Federated-Learning-for-Misbehaviour-Detection-with-Variational-Autoencoder-and-Gaussian-Mixture-Mode}}.

\begin{table}[]
\centering
\begin{tabular}{ll}
\hline
\textbf{Framework} & Flower            \\ \hline
\textbf{Model}     & VAE / AE \\ \hline
\textbf{Aggregation function}     & Fed+ \\ \hline
\textbf{Clients}   & From Fig. \ref{fig:value_counts}         \\ \hline
\textbf{Rounds}    & 30                \\ \hline
\textbf{Epochs}    & 1                 \\ \hline
\end{tabular}
\caption{Summary of parameter of our federated scenarios.}
\label{tab:summary_fl}
\end{table}

\subsection{Client division}
\begin{figure} [!ht]
	\centering
		\includegraphics[width=\columnwidth]{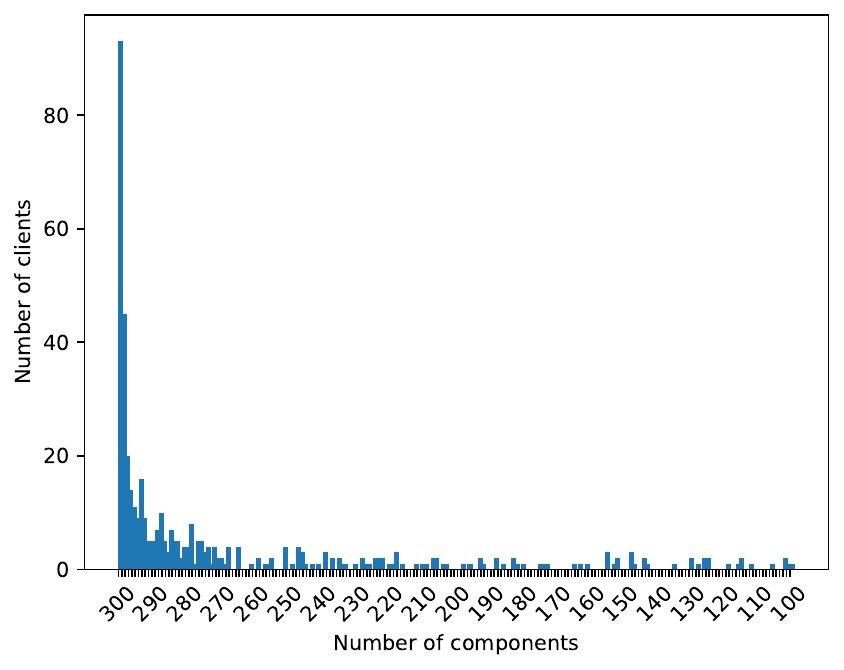}
	\caption{Number of clients per cluster depending on the number of components}
	\label{fig:value_counts}
\end{figure}

The use of GMMs requires the specification of a certain number of components (clusters) before training the model. Furthermore, such number will also determine the input size of the AE/VAE. Instead of selecting a fixed number of components for all the clients, we compute the optimal number of components for each client using silhouette analysis as stated in Section \ref{sec:System}. Consequently, clients are grouped based on their optimal number of components since different networks with different input sizes will be created (one per group). In order to apply the GMM, we have to specify the type of the covariance matrix. In our case, we choose the covariance type which fits better to our model, which is diagonal.  Fig. \ref{fig:value_counts} shows the number of clients/cars that resulted in having a certain number of components as optimal. As shown, the most common number of components was 300. Hence, we will have as many federated scenarios as groups of cars are created based on their optimal number of components.

As will be described in Section \ref{sec:particular_case}, we provide an extensive analysis for the group of clients whose optimal number of GMM components is 298. Such analysis is also applicable for each group of clients with the same number of components. In this sense, in order to avoid repeating the process 104 times (as it is the total number of groups), we also provide the final results for the rest of the groups with 5 or more clients (Section \ref{sec:general_case}). The reason for choosing this grouping is to display a better picture comparing all clients' performance. Although the group with 298 components has fewer clients compared to the clusters of 300 and 299 components (20, 93, and 45 clients respectively), the clients' average sample size is similar between those groups (20453,  19983, and 21004 respectively), as well as the standard deviation (7677, 9760, and 10885). Therefore, the clients of this particular case have a similar distribution, and having a small number of clients can ease results' interpretation and plotting. 

Moreover, our analysis includes a comparison of the performance regarding the autoencoder process depending on the learning rate (\textit{lr}) using an AE or VAE. It should be noted that both AE and VAE are built as similarly as possible to ease their comparison. Specifically, the input layer's size is the number of GMM components. In the encoder, the first hidden layer is half the size of the input layer. Then, the AE's latent space and the VAE's sample layer are third of the size of the input layer. Finally, the decoder has two layers: the first one is half of the size of the input layer, and the output layer has the size of the input layer. All these layers use ReLU as the activation function except the last one, which employs the sigmoid function, and both use RMSprop as the optimizer. For the rest of the clusters, we compare the results considering AE and VAE with the best \textit{lr}. 

\subsection{Particular case analysis: Clients with 298 components}\label{sec:particular_case}
As shown in Fig. \ref{fig:value_counts}, 20 clients are grouped with 298 components. In this section, we analyse the impact of choosing between AE or VAE in the federated setting, and we also compare the results with the distributed scenario where clients only train locally, i.e., they do not share any result derived from such training with the server. It should be noted that we distinguish between AE/VAE accuracy (autoencoder part), the accuracy obtain during the application of AE/VAE in phase 3 in \ref{sec:System}, and the general model's accuracy (total accuracy), which is the accuracy obtained based on the use of GMM and AE/VAE. The reason is that GMM process is nearly the same every round as it is not part of the federated training, so we analyse the impact of the AE/VAE model in the performance of the model. 

\begin{figure} [!ht]
	\centering
		\includegraphics[width=\columnwidth]{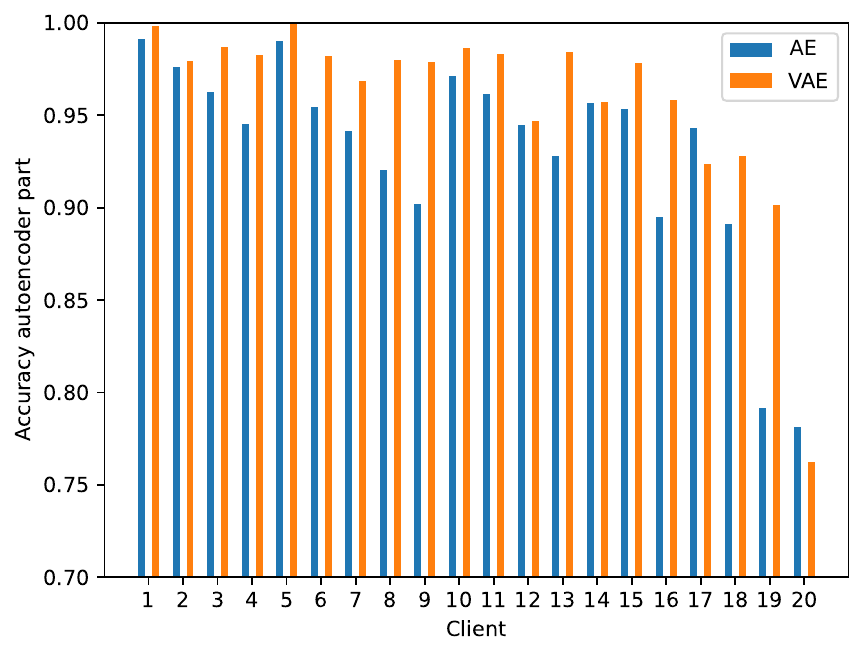}
	\caption{Comparison of accuracy of autoencoder part between best case of VAE ($lr=0.05$) and AE ($lr=0.005$) in each client of the  298-component cluster}
	\label{fig:comp_client}
\end{figure}

\begin{figure} [!ht]
	\centering
		\includegraphics[width=\columnwidth]{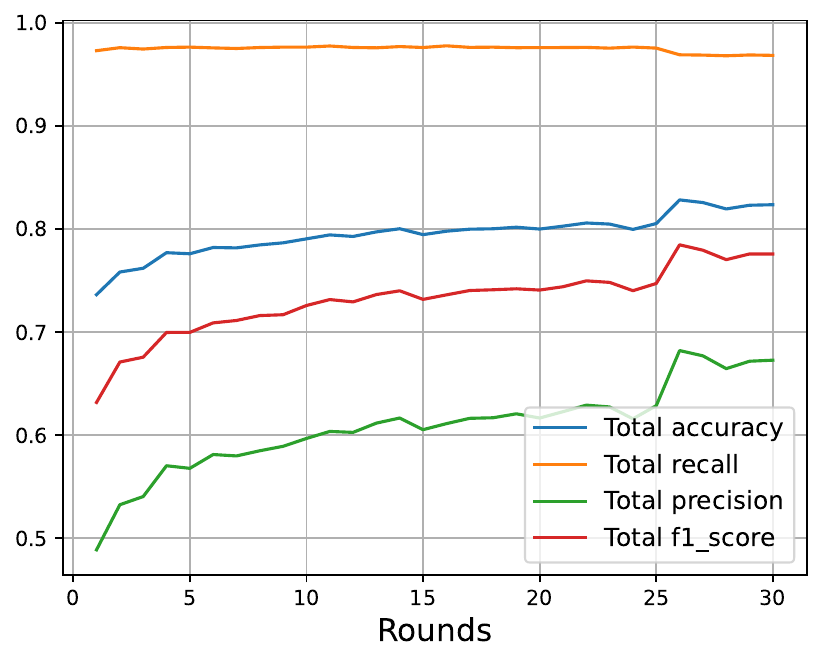}
	\caption{Metrics of the federated scenario using GMM and VAE of the  298-component cluster}
	\label{fig:metrics_GMMVAE}
\end{figure}

Table \ref{tab:acc_lr} shows the accuracy values of the autoencoder part (that is, using AE/VAE) considering different \textit{lr}, marking in bold the best \textit{lr} for each method. According to the results, we best value obtained is for VAE when $lr=0.05$. In particular, considering the best case for each setting, the federated VAE reaches a value of 0.972 (\textit{lr} = 0.05), whereas the distributed VAE, federated AE, and distributed AE reach 0.928 (\textit{lr} = 0.001), 0.923 (\textit{lr} = 0.005), and 0.936 (\textit{lr}= 0.005) respectively. Comparing the performance in particular of the best cases of federated VAE and federated AE, in Fig. \ref{fig:comp_client}, we compare the accuracy of each client specifically. In this figure, we see that in almost all clients, the VAE reaches better performance than the AE. Having set the best \textit{lr} for the VAE, Fig. \ref{fig:metrics_GMMVAE} shows the final metrics of the general model, that is, considering the metrics obtained by applying GMM and the VAE. The accuracy of the model is 0.824, and the recall, precision, and f1-score are 0.968, 0.672, and 0.775 respectively.


\begin{table*}[]
\centering
\begin{tabular}{lllll}
\textbf{Learning rate} & \textbf{Federated VAE} & \textbf{Distributed VAE} & \textbf{Federated AE} & \textbf{Distributed AE} \\ \hline
0.001                  & 0.920                & \textbf{0.928}                  & 0.90             & 0.921                  \\ \hline
0.005                  & 0.892                & 0.915                   & \textbf{0.923}      & \textbf{0.936 }                  \\ \hline
0.01                   & 0.767                & 0.792                   & 0.893               & 0.911                  \\ \hline
0.05                   & \textbf{0.972}       & 0.853                   & 0.455               & 0.3640                  \\ \hline
0.1                    & 0.953                & 0.901                   & 0.457               & 0.447                  \\ \hline
\end{tabular}
\caption{Accuracy of the autoencoder part of different methods varying the \textit{lr} in the  298-component cluster}
\label{tab:acc_lr}
\end{table*}

In order to justify the grid of GMM components chosen in Section \ref{sec:System}, Table \ref{tab:timeComparison} provides a comparison of the time consumed by the principal processes of our approach depending on the number of the GMM components and the accuracy achieved. As the number of components increases, the time required is higher, especially because of the silhouette analysis. Nevertheless, although the accuracy also grows, the increase of each step is getting lower, from 0.62 to 0.824. Hence, we set the maximum number of components of the grid at 300, since a wider range will consume an enormous amount of time for a reduced improvement in terms of accuracy. 


\begin{table}[]
\begin{tabular}{l|llll}
\textbf{Number of components} & \textbf{10} & \textbf{100} & \textbf{200} & \textbf{300} \\ \hline
\textbf{Silhouette analysis}  & 947         & 21357        & 54298        & 110132       \\ \hline
\textbf{Training GMM}         & 15          & 50           & 202          & 236          \\ \hline
\textbf{Training RBM}          & 28          & 286          & 1028         & 2122 \\ \hline
\textbf{Accuracy}          & 0.62          & 0.705          & 0.783         & 0.824 \\ \hline
\end{tabular}
\caption{Time comparison (in seconds) of the main processes of our approach depending on the number of GMM components, and accuracy}
\label{tab:timeComparison}
\end{table}

Finally, we also compare our approach with a scenario where the GMM and histograms are not used. In this case, the VAE is applied as a baseline employing the benign dataset, so potential misbehavior is detected using the reconstruction error. This method reaches 0.587 of accuracy, which supports the need to use the combination of GMM and histograms with the VAE.

\subsection{General case: clusters with 5 or more clients}\label{sec:general_case}
\begin{figure} [!ht]
	\centering
		\includegraphics[width=\columnwidth]{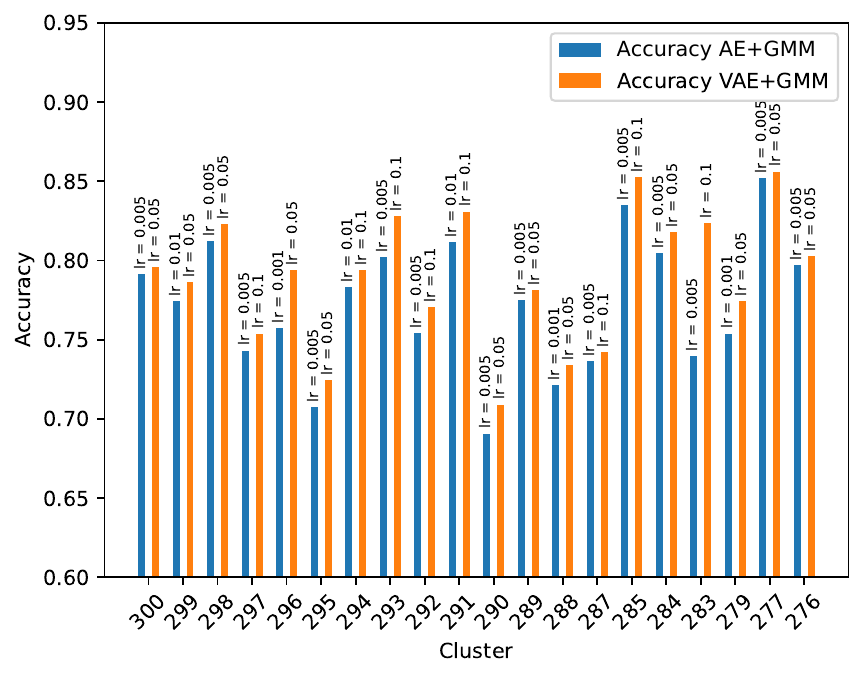}
	\caption{Comparison between accuracy of AE+GMM and VAE+GMM in all clusters}
	\label{fig:resto_clientes_total}
\end{figure}

\begin{figure} [!ht]
	\centering
		\includegraphics[width=\columnwidth]{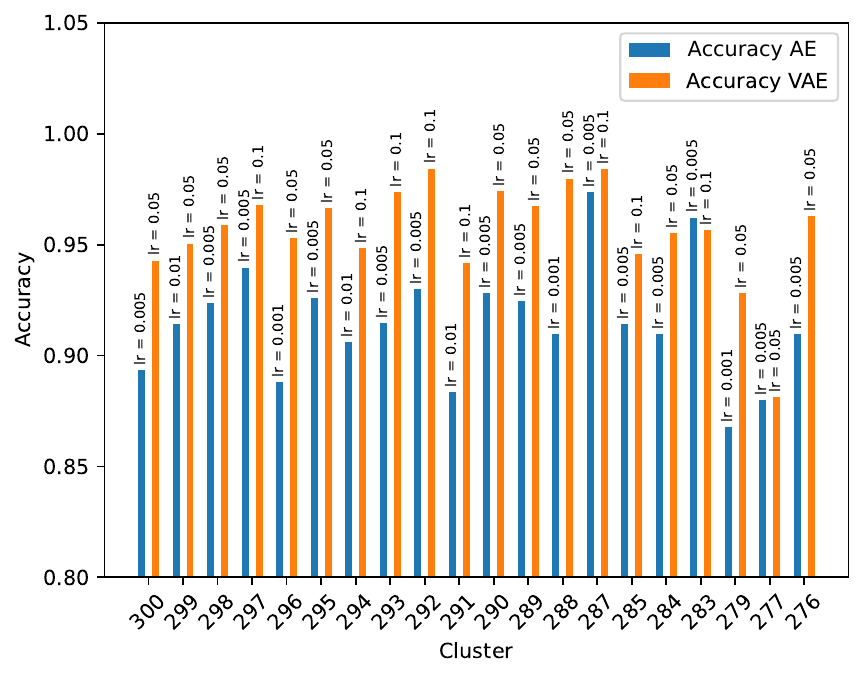}
	\caption{Comparison between accuracy of AE and VAE in all clusters}
	\label{fig:resto_clientes:ae}
\end{figure}

This section shows the results for all clusters (any number of components) of Fig. \ref{fig:value_counts} that contain 5 or more clients. Fig. \ref{fig:resto_clientes_total} and Fig. \ref{fig:resto_clientes:ae} provide the accuracy of the general model and the autoencoder part respectively, comparing when AE or VAE is used. For each case, we select the best \textit{lr}.
As shown, the results using the VAE are higher compared to AE for almost every cluster.

\begin{figure} [!ht]
	\centering
		\includegraphics[width=\columnwidth]{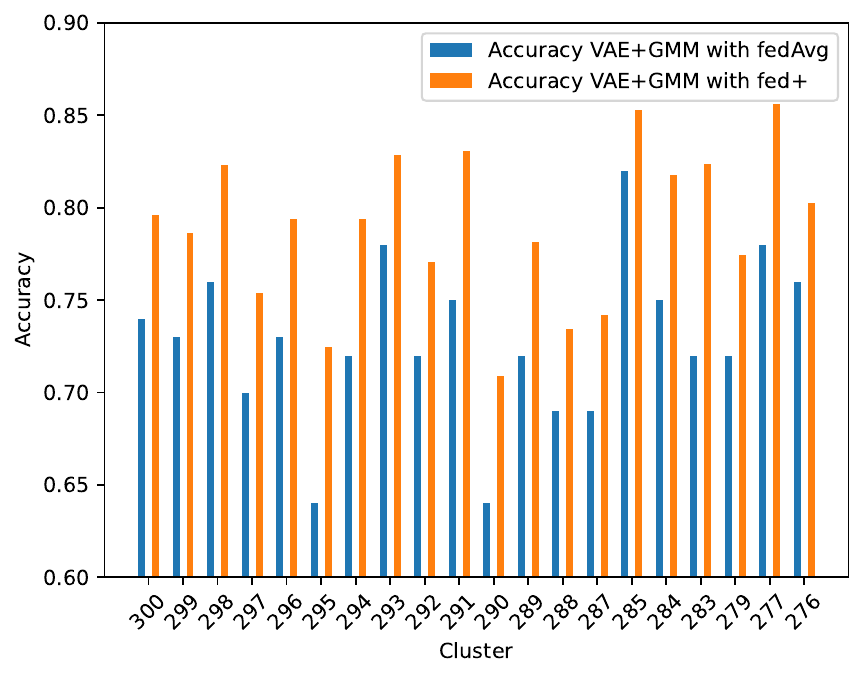}
	\caption{Comparison between total accuracy of VAE+GMM using Fed+ of FedAvg in all clusters}
	\label{fig:fedpvsfedavgTotal}
\end{figure}

\begin{figure} [!ht]
	\centering
		\includegraphics[width=\columnwidth]{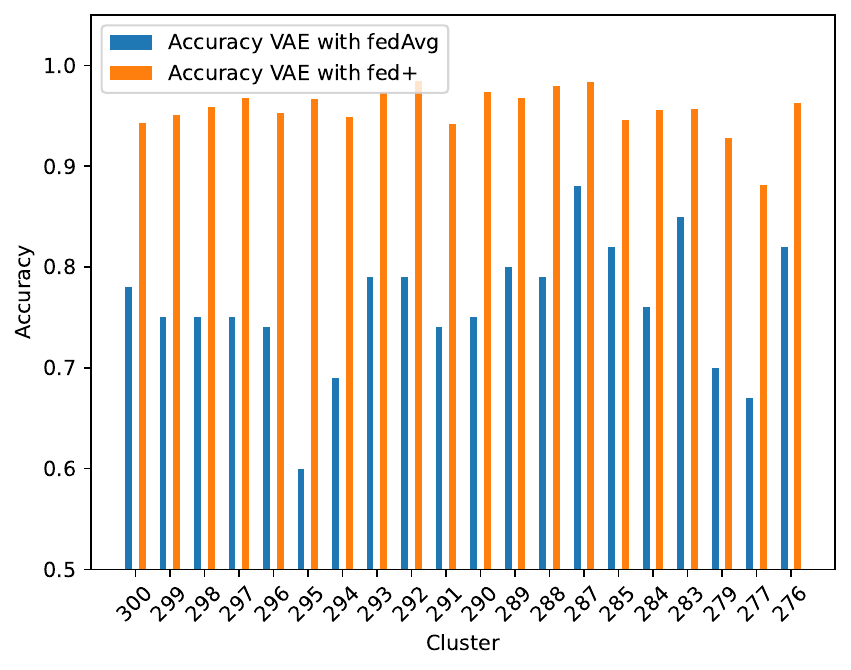}
	\caption{Comparison between total accuracy of VAE using Fed+ or FedAvg in all clusters}
	\label{fig:fedpvsfedavgSolo}
\end{figure}

For the sake of completeness, we show the comparison of our method using FedAvg as aggregation function in phase 2 mentioned in section \ref{sec:Preeliminaries}. In Fig. \ref{fig:fedpvsfedavgTotal} and \ref{fig:fedpvsfedavgSolo} show the total accuracy of the VAE and GMM, and the accuracy of the VAE using either Fed+ or FedAvg. The \textit{lr} used in these figures are the same of the ones used in Fig. \ref{fig:resto_clientes_total} and \ref{fig:resto_clientes:ae}. In these pictures, we can clearly appreciate that in all clusters FedAvg achieves worse performance than the case with Fed+.

Furthermore, we compare our evaluation results with recent works on the same dataset. In particular, \cite{moulahi2022privacy} uses several supervised ML algorithms, including Support Vector Machines (SVM) K-Nearest Neighbour (KNN), Naïve Bayes, and Random Forest. In general, we achieve similar results but with clearly higher values for certain metrics, such as recall and f1-score in which they achieve 0.5 and 0.6 respectively, whilst we get 0.96 and 0.77, respectively. Furthermore, it should be noted that the mentioned work achieves those results using supervised learning approaches, and an artificial division of the VeReMi dataset, which is split into five equally balanced clients, an iid division that not reflect the real-case scenarios. In our approach, clients' data are based on the vehicles in the VeReMi dataset, so that each vehicle only trains with its local data. In this sense, our division corresponds a more realistic scenario, which is characterized by non-iid data distributions. We also compare our unsupervised FL approach with the case of using a supervised MLP, which provides an accuracy value of 0.88. It should be noted that even in this case with non-iid data, the accuracy value is higher compared with the work previously mentioned. The main reason is that we mitigate such aspect by using SMOTE-Tomek to obtain a more balanced dataset, and Fed+ as aggregation function.



\section{Conclusions and future work}\label{sec:Conlutions}



This work proposed an unsupervised FL approach for misbehavior detection in vehicular environments based on the use of GMM and VAE. The resulting system was exhaustively evaluated considering a realistic scenario where each FL client is intended to train with its local data. Unlike most of existing approaches, we deal with non-iid data distributions and convergence issues by considering dataset balancing techniques, as well as alternative aggregation functions. Our evaluation also measures the impact of the learning rate on the federated training process, and compare the approach with recent works using the VeReMi dataset. According to the results, our proposed system provides a performance close to supervised approaches, which require labelled datasets for training. As part of our future work, we plan to build VAEs to identify different classes of misbehavior, that is, beyond binary classification. Furthermore, as the choice of the learning rate could have a significant impact on the performance, we will design an approach to dynamically set the learning rate throughout the training rounds. Additionally, we also plan to analyze the possibility of dynamically selecting specific clients to reduce the bandwidth required during the federated training process.


\section*{Acknowledgements}
This work has been sponsored by the EC through H2020 ERATOSTHENES (g.a. 101020416) and the European Union’s Horizon Europe research and innovation funding programme under the Marie Skłodowska-Curie grant agreement No 101065524. It also forms part of the ThinkInAzul programme and was supported by MCIN with funding from European Union NextGenerationEU (PRTR-C17.I1) by Comunidad Autónoma de la Región de Murcia - Fundación Séneca as well as by the HORIZON-MSCA-2021-SE-01-01 project Cloudstars (g.a. 101086248) and ONOFRE Project PID2020-112675RB-C44 funded by MCIN/AEI/10.13039/501100011033.

\small
\bibliographystyle{IEEEtran}
\bibliography{biblio}

\end{document}